\documentclass{article}
\usepackage{spconf,amsmath,graphicx}
\usepackage[OT1]{fontenc}
\usepackage{subfigure}
\usepackage{multirow}
\usepackage{multicol}
\usepackage{booktabs, tabularx}
\usepackage{graphicx}
\usepackage{algorithm}
\usepackage{algorithmic}
\usepackage{xcolor}
\title{Solving The Long-Tailed Problem via Intra- and Inter-Category Balance}
\name{Renhui Zhang, Tiancheng Lin, Rui Zhang, Yi Xu}
\address{Shanghai Jiao Tong University}
\begin{document}
%\ninept
%
\maketitle
\begin{abstract}
Benchmark datasets for visual recognition assume that data is uniformly distributed, while real-world datasets obey long-tailed distribution. 
Current approaches handle the long-tailed problem to transform the long-tailed dataset to uniform distribution by re-sampling or re-weighting strategies.
These approaches emphasize the tail classes but ignore the hard examples in head classes, which result in performance degradation.
In this paper, we propose a novel gradient harmonized mechanism with category-wise adaptive precision to decouple the difficulty and sample size imbalance in the long-tailed problem, which are correspondingly solved via intra- and inter-category balance strategies.
Specifically, intra-category balance focuses on the hard examples in each category to optimize the decision boundary, while inter-category balance aims to correct the shift of decision boundary by taking each category as a unit. 
Extensive experiments demonstrate that the proposed method consistently outperforms other approaches on all the datasets.
\end{abstract}
\begin{keywords}
long-tailed distribution, image classification, adaptive precision, gradients harmonizing
\end{keywords}
\section{Introduction}
\label{sec:intro}
% \lin{Long-tailed distribution is a common phenomenon in the real world \cite{van2018inaturalist}.
% In a deep learning task, it appears that most samples in the dataset belong to a few 
% dominant categories (head classes), while a large number of other categories 
% (tail classes) claim very few samples \cite{van2017devil, liu2019large}. 
% However, the most popular benchmark datasets for deep learning tasks of visual recognition 
% are nearly uniformly distributed in both training and testing data, such as CIFAR \cite{krizhevsky2009learning, torralba200880} and ImageNet ILSVRC 2012 \cite{deng2009imagenet, russakovsky2015imagenet}. As a result, standard methods designed for benchmarks perform poorly on long-tailed problems in that training and testing data are no longer IID (identical \& independent distribution). Specifically, when training on long-tailed data, the extreme imbalance between head and tail classes leads to the decision boundary biasing towards dominant classes. }

\begin{figure}[t]
    \centering
    \subfigure[No balance.]{
        \includegraphics[width=0.3\linewidth]{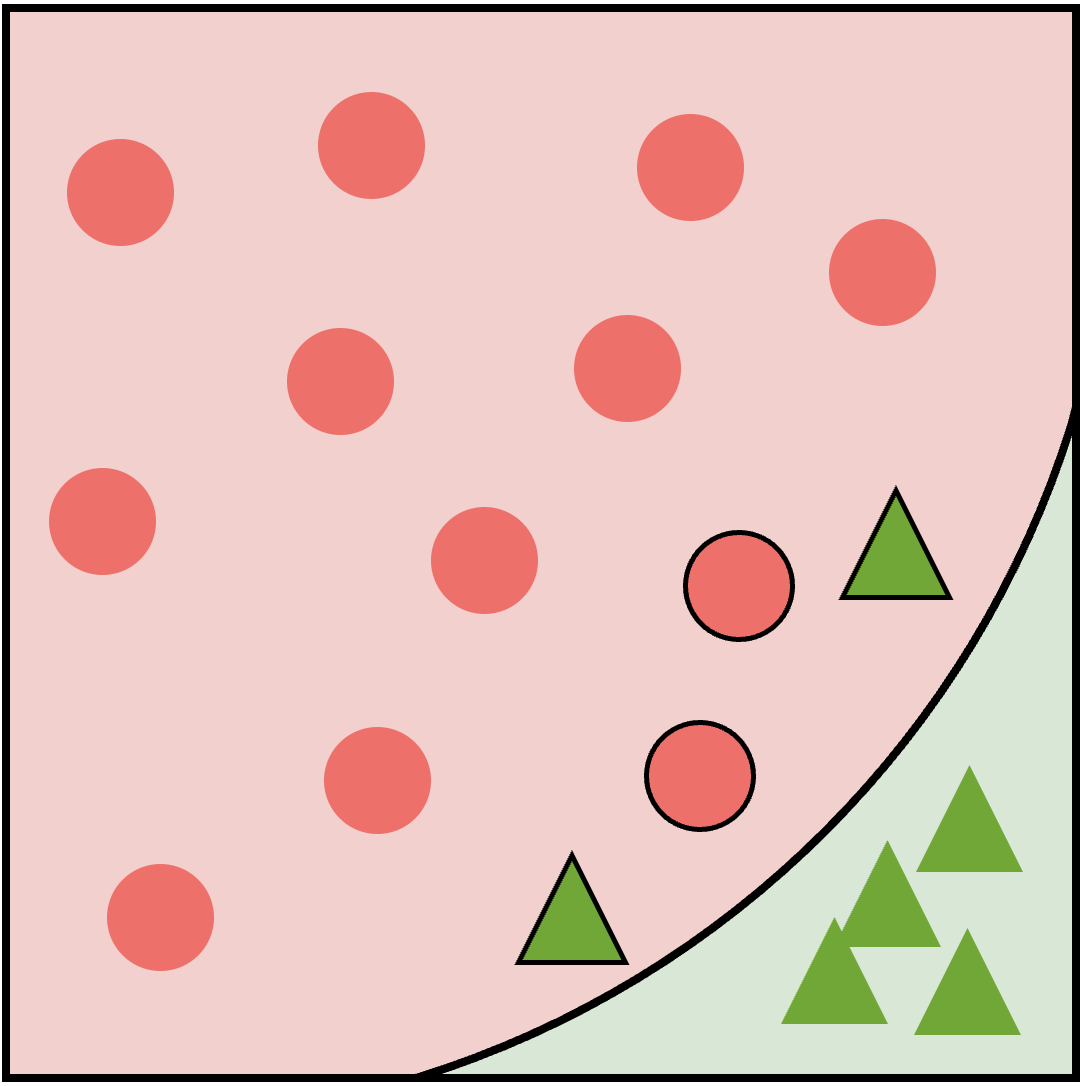}
        \label{fig:ori}}
    \subfigure[Previous works.]{
        \includegraphics[width=0.3\linewidth]{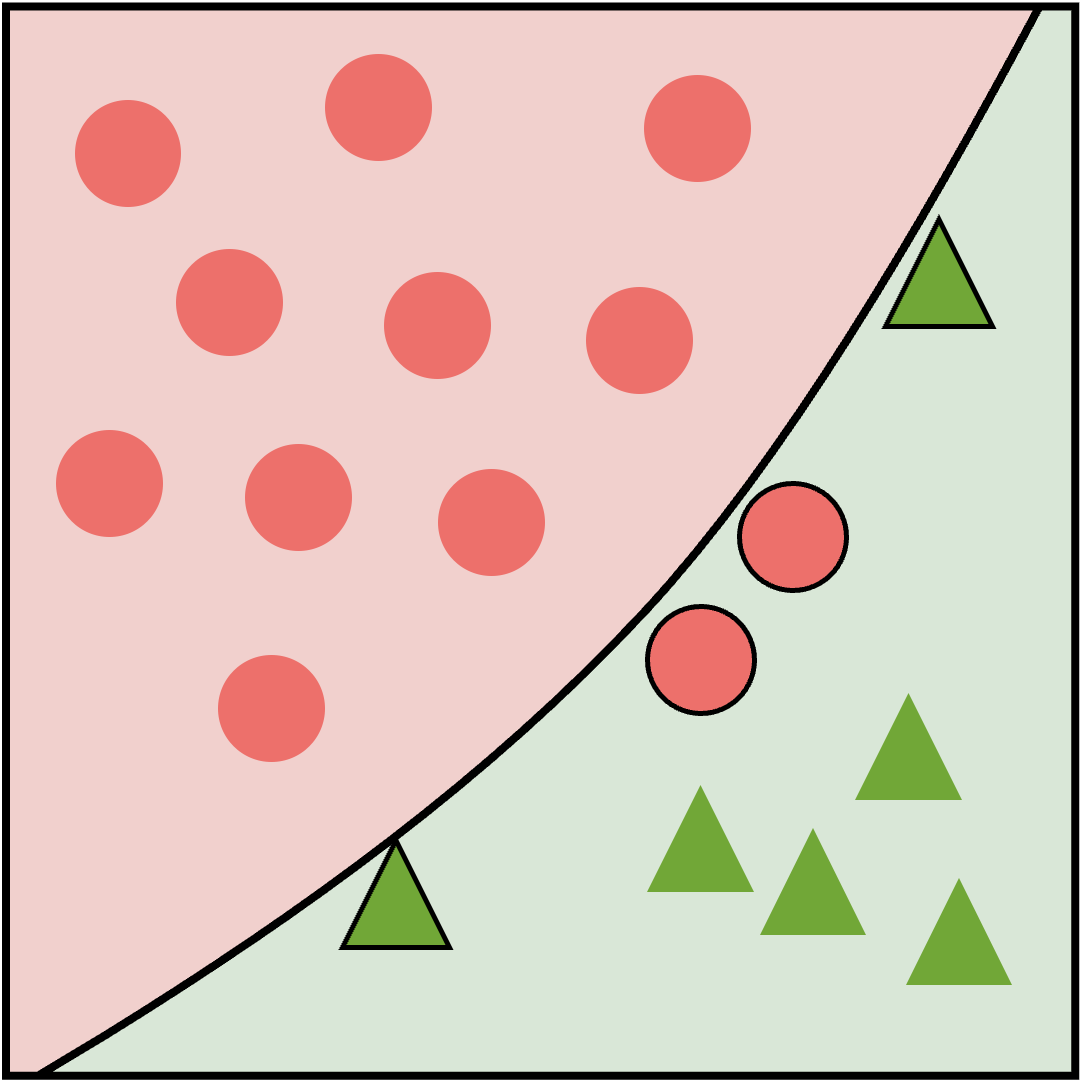}
        \label{fig:inter}}
    \subfigure[Ours.]{
        \includegraphics[width=0.3\linewidth]{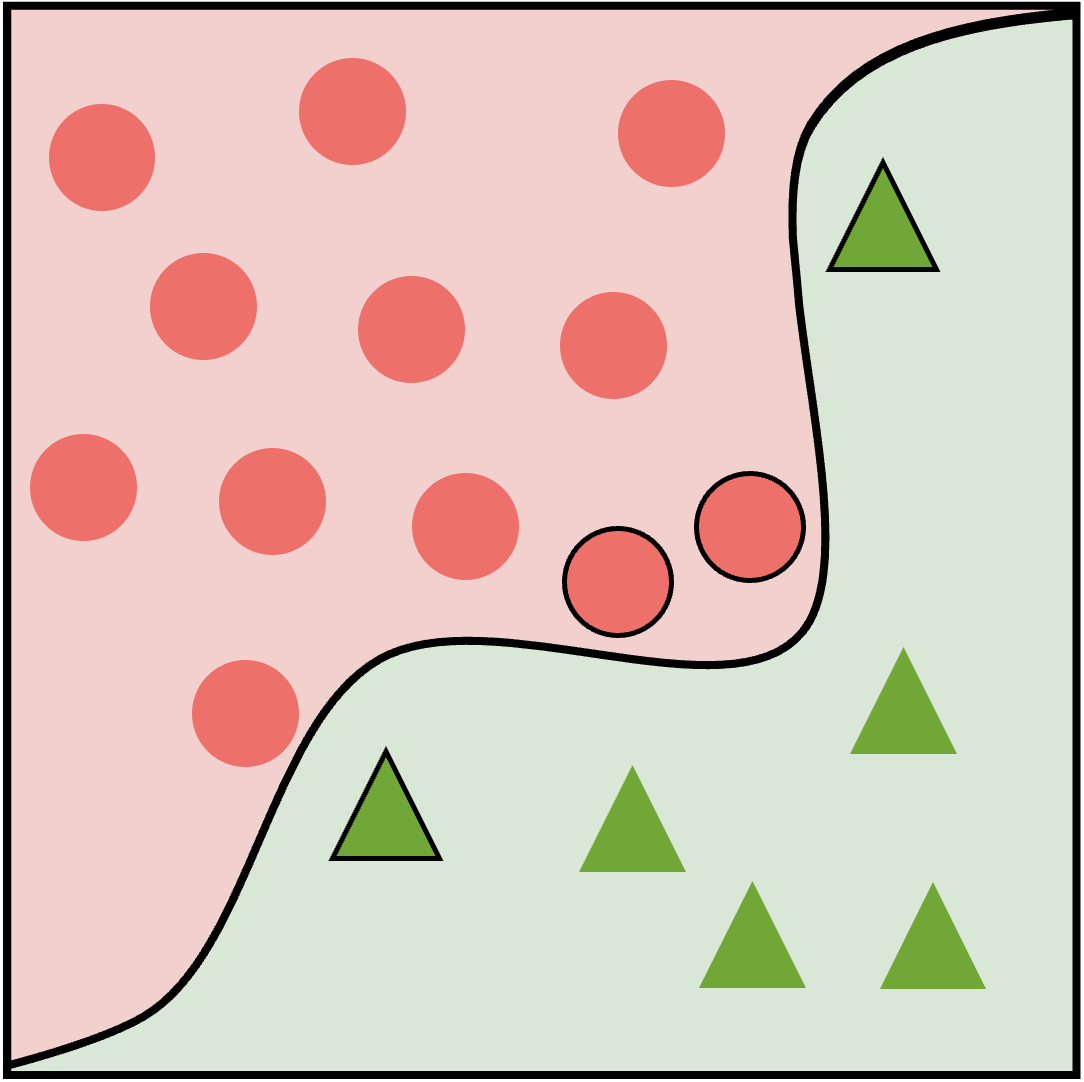}
        \label{fig:inter_ra}}
    \caption{Red circles and green triangles denote examples from head class and tail class respectively, and hard examples are annotated with black boarder.
    (a) Decision boundary (the black line) bias towards head classes in long-tailed dataset. 
    (b) Previous works attempt to correct the bias by inter-category balance but ignores the hard examples (red circles or green triangles with black border).
    (c) Our method corrects the bias problem through emphasizing both hard examples and tail-class by intra- and inter-category balance.}
    \label{fig:longtail}
\end{figure}

Long-tailed distribution is a common phenomenon in the real world \cite{van2018inaturalist}.
In a deep learning task, most examples in the dataset belong to a few 
dominant categories (head classes), while a large number of other categories 
(tail classes) claim very few examples \cite{van2017devil, liu2019large}. 
In fact, the most popular benchmark datasets for deep learning tasks of visual recognition 
are nearly uniformly distributed , such as CIFAR \cite{krizhevsky2009learning, torralba200880} and ImageNet ILSVRC 2012 \cite{deng2009imagenet, russakovsky2015imagenet}.
% As a result, models designed for these datasets, such as ResNet \cite{he2016deep}, perform poorly on long-tailed datasets especially for tail classes \cite{buda2018systematic}. 
% As shown in figure \ref{fig:ori}, the extreme imbalance between head and tail classes leads to the decision boundary biasing towards dominant classes (the black solid line). 
% \lin{It is not the fault of model. My version: }
As a result, standard methods designed for benchmarks perform poorly on long-tailed problems \cite{buda2018systematic} in that the extreme imbalance between head and tail classes leads to the decision boundary bias towards head classes, 
as shown in Fig.~\ref{fig:ori}.
In order to remove the bias, current approaches deal with the long-tailed problem with an attempt to transform the long-tailed dataset to uniform distribution by re-sampling  \cite{kubat1997addressing, chawla2002smote, chu2020feature, jiawei2020balanced, yin2018feature} or re-weighting strategies~\cite{cao_learning_2019, jamal_rethinking_2020, fan2017learning, iranmehr2019cost}.
% To pay more attention to the samples of small quantity, they emphasized the tail classes and alter the decision boundaries by introducing inter-category balance. 
% \lin{
In particular, they assign different weights (based on effectiveness~\cite{cui2019class} or inverting data presenting frequency~\cite{tan2020equalization}), margins (symmetric~\cite{iranmehr2019cost} or asymmetric~\cite{wang2021seesaw}) and sampling strategies(over-sampling~\cite{chawla2002smote} or under-sampling~\cite{kubat1997addressing}) to the examples from different categories, which is called inter-category balance in this paper.
% }
% They emphasized the tail classes and alter the decision boundary by inter-category balance. 
% However, the imbalance between easy and hard samples is ignored in these studies, the imbalance in difficulty implies that the contributions of easy and hard samples are different.
% However, the imbalance between easy and hard samples is ignored in these studies.
% Therefore, a few members in head classes cannot be well classified due to their very limited quantities and high ambiguities. 
However, most of the previous works increase the accuracy of tail classes at the cost of the head classes.
It is mainly caused by the under-fitting of hard examples in head classes.
% As shown in Fig~\ref{fig:inter}, previous work attempts to correct the bias by inter-category balance but ignores the hard samples, which result in the performance cost on head classes.
As shown in Fig.~\ref{fig:inter}, when the decision boundary of the tail class is expanded, the hard examples near the decision boundaries are more likely to be misclassified than the easy counterparts in the head class. 

% \begin{figure}
%   \centering
%   \includegraphics[width=0.8\linewidth]{figs/mainfigv4.pdf}
%   \caption{Previous methods deal with longtail problem applies inter-category balance (blue arrow and dotted lines) only. Besides intra-category balance (red/green arrow and dotted cycles), our method indicates that intra-category balance optimizes the feature space while inter-category balance  corrects the shift of decision boundary. \lin{The left part is unnecessary, maybe follow `Class-Balanced Loss Based on Effective Number of Samples'. If you mention hard examples and outliers, please show them in Fig 1.}}
%   \label{fig:longtail}
% \end{figure}
In this work, we point out that the imbalance in long-tailed problems can be summarized as the imbalance in difficulty and sample size. 
The former one implies that the contributions of easy and hard examples should be different to the model learning process
while the latter leads to the bias of the decision boundary.
% In this work, we address that the imbalance in long-tailed problem can be 
% summarized to the imbalance in sample size and the imbalance in difficulty.
% The imbalance in sample size leads to the bias of the decision boundary.
% And the imbalance in difficulty implies that the contributions of easy and hard samples are different.
% Inspired by the analysis of the two types of imbalance, in this paper we solve 
% the long-tailed problem by introducing strategies of intra-category balance based on adaptive precision 
% gradient norm histogram and 
% inter-category balance based on the relative margin. \lin{double check.} 
% Figure~\ref{fig:longtail} shows the schematic of our method, 
% intra-category balance aims to obtain an optimal feature space by emphasizing hard samples.
% Inter-category balance aims to correct the shift of decision boundary as the blue arrow denotes.
% Integrating intra- and inter-category balance into our method, we can obtain a better decision boundary as compared with other method. 
% Inspired by the analysis of the two types of imbalance, 
Accordingly,
we propose a novel
% \textit{adaptive gradient harmonized mechanism} 
% {\color{red}
gradient harmonized mechanism with category-wise adaptive precision (GHM-CWAP) to decouple the difﬁculty and sample size.
Correspondingly, intra- and inter-category balance strategies are introduced to calculate the weight of each example, which tackle the imbalance of both difﬁculty and sample size respectively.
% }
% to decouple the difficulty and sample size.
% Based on the difficulty and sample size,
% intra- and inter-category balance strategies are introduced, where weight of each example is decided by both sample size and difficulty. 
Specifically, intra-category balance seeks to rectify the decision boundary by focusing on the hard examples
while
inter-category balance aims to correct the shift of decision boundary by taking all examples of one category as a unity. 
By rationally integrating two balance strategies, we can obtain a better decision boundary as illustrated in Fig.~\ref{fig:inter_ra}.
% inter-category balance aims to harmonize the contribution of head and tail classes (the blue arrow).
% Hence the bias of decision boundary would be corrected, as the blue dashed line.
% Intra-category balance aims to obtain a optimal decision space by emphasizing hard samples.
% Hard samples are distinguished by the gradient of cost function.
% delete

Extensive experiments show that the proposed method achieves the comprehensive balance for the long-tailed problem and consistently outperforms other approaches on all datasets
without sacrificing the head classes.

\section{Related Work}
\textbf{Intra-category balance} aims to harmonize the contributions of hard and easy examples.
Previous works, which study example difficulty in terms of loss, assign higher weights to hard examples~\cite{malisiewicz2011ensemble, dong2017class, lin2017focal, li2019gradient}.
% Hard mining method, which is a kind of intra-category balance strategy, is first introduced to solve the imbalance problem of hard and easy samples in object detection. \lin{double check}
% Intra-category balance is first introduced to solve 
% the imbalance problem of hard and easy samples in object detection.
% Focal loss \cite{lin2017focal}, which attempts to 
% balance the contributions of hard and easy samples,
% assigning weight to samples according to the gradient of loss function.
% Furthermore, GHM loss \cite{li2019gradient} is proposed to 
% harmonize the contributions of hard, easy samples.
% Considering that the quantity of hard samples is less than easy samples and outliers.
% GHM constructs a gradient norm histogram and reweight the gradient of each sample according to inverse proportion of its frequency of occurrence.
% It should be noted that above methods are not suitable for long-tail problem, in which the gradient norm histogram is dominated by head classes while the gradients of tail classes would be submerged.
% \lin{
Since hard examples are much less than easy examples,
these methods couple the sample size and difficulty, 
which result in that examples from head classes tend to be regarded as easy. 
% The main problem of these method is that 
Therefore, directly applying these methods to long-tailed problem is unsuitable, as the hard example mining process may be dominated by head classes.
% }
% GHM constructs a gradient norm histogram and 
% reweight the the gradient of each sample 
% according to inverse proportion of the gradient norm distribution.
% Since the quantity of hard samples is less than easy samples and outliers 
% in the gradient norm histogram, hard samples would be emphasized 
% while easy samples and outliers are weakened.
% Above methods are not suitable for long-tail problem 
% as the gradient norm histogram would be dominated by head classes 
% while the gradients of tail classes would be submerged.

\noindent\textbf{Inter-category balance strategy} is commonly used in imbalance problem, 
and mainly contains two ways, 
re-sampling \cite{kubat1997addressing, chawla2002smote, chu2020feature, jiawei2020balanced, yin2018feature} and 
re-weighting \cite{cao_learning_2019, jamal_rethinking_2020, fan2017learning, iranmehr2019cost}.
Re-sampling aims to balance the sample size of each category used for training 
by data or feature augmentation.
% In contrast, re-weighting aims to harmonize the contribution of samples 
% during the training phase, assigning various weight to different samples 
% or adding various margin to the model output logits. 
In contrast, re-weighting aims to harmonize the contribution of examples during training, assigning weight to examples or adding margin to the model output logits adaptively. 
Above methods attempt to convert the long-tailed dataset to 
uniform distribution by emphasizing the tail classes. 
However, they ignore the imbalance of difﬁculty among the examples during learning process, which result in the performance degradation on head classes.

\begin{table*}[!htb]
  \caption{Top-1 accuracy (\%) of ResNet-32 with various loss function on long-tailed CIFAR-10/100 and TinyImageNet. 
  Imbalance facotr means the ratio of sample size of head classes to tail classes.}
  % Class Balanced means the loss is re-weighted by the reciprocal of samples size.}
  \label{tab:big_table}
  \resizebox{\textwidth}{!}{
    \begin{tabular}{l|cccc|cccc|cccc}
      \toprule
      Dataset          & \multicolumn{4}{c|}{Long-Tailed CIFAR-10} & \multicolumn{4}{c|}{Long-Tailed CIFAR-100} & \multicolumn{4}{c}{Long-Tailed TinyImagenet} \\ \hline
      Imbalance factor       & 500      & 100      & 10       & 1        & 500       & 100      & 10       & 1        & 500       & 100       & 10        & 1        \\ \midrule
      Softmax          & 59.76    & 71.74    & 86.69    & 93.00    & 30.90     & 38.77    & 57.21    & 71.00    & 31.42     & 39.06     & 54.51     & \textbf{63.32}    \\
      Class Balanced   & 58.50   & 73.82    & 87.18   & 92.71    & 31.23     & 39.12    & 56.45    & 70.75    & 30.90     & 39.48     & 54.17     & 63.06    \\
      Focal \cite{lin2017focal}            & 57.72    & 72.43    & 86.70    & 92.66    & 30.31     & 38.12    & 56.12    & 70.13    & 30.93     & 39.23     & 54.00     & 63.22    \\
      GHM \cite{li2019gradient}             & 41.50    & 51.66    & 67.87    & 79.35    & 12.85     & 14.02    & 24.49    & 34.04    & 8.00      & 8.03      & 9.92      & 10.13     \\
      Effective Number \cite{cui2019class} & 57.97    & 71.47    & 87.13    & 92.98    & 30.40     & 39.32   & 57.37    & 70.75    & 30.90     & 39.48     & 54.17     & 63.06   \\
      Equalization \cite{tan2020equalization}             & 58.29    & 72.61    & 87.08    & \textbf{93.04}  & 30.82     & 39.84    & 58.14    & \textbf{71.66}    & 29.63     & 37.39     & 52.75     & 62.84    \\
      Seesaw \cite{wang2021seesaw}           & 64.48    & 75.18    & \textbf{87.76}    & 93.03    & 33.63     & 40.87    & \textbf{57.83}    & 71.35    & 34.07     & 40.98     & 54.60     & 63.05    \\

      ours             & \textbf{66.13}    & \textbf{75.39}    & 87.29    & 92.75    & \textbf{33.85}     & \textbf{41.59}    & 57.81    & 71.41    & \textbf{34.79}     & \textbf{42.66}     & \textbf{55.35}     & 63.21   \\ \bottomrule
    \end{tabular}
  }
\end{table*}

\section{Method}

\begin{figure}
  \begin{minipage}[b]{0.65\linewidth}
    \subfigure[Normalized gradient norm distributions.]{
      \includegraphics[width=0.98\linewidth]{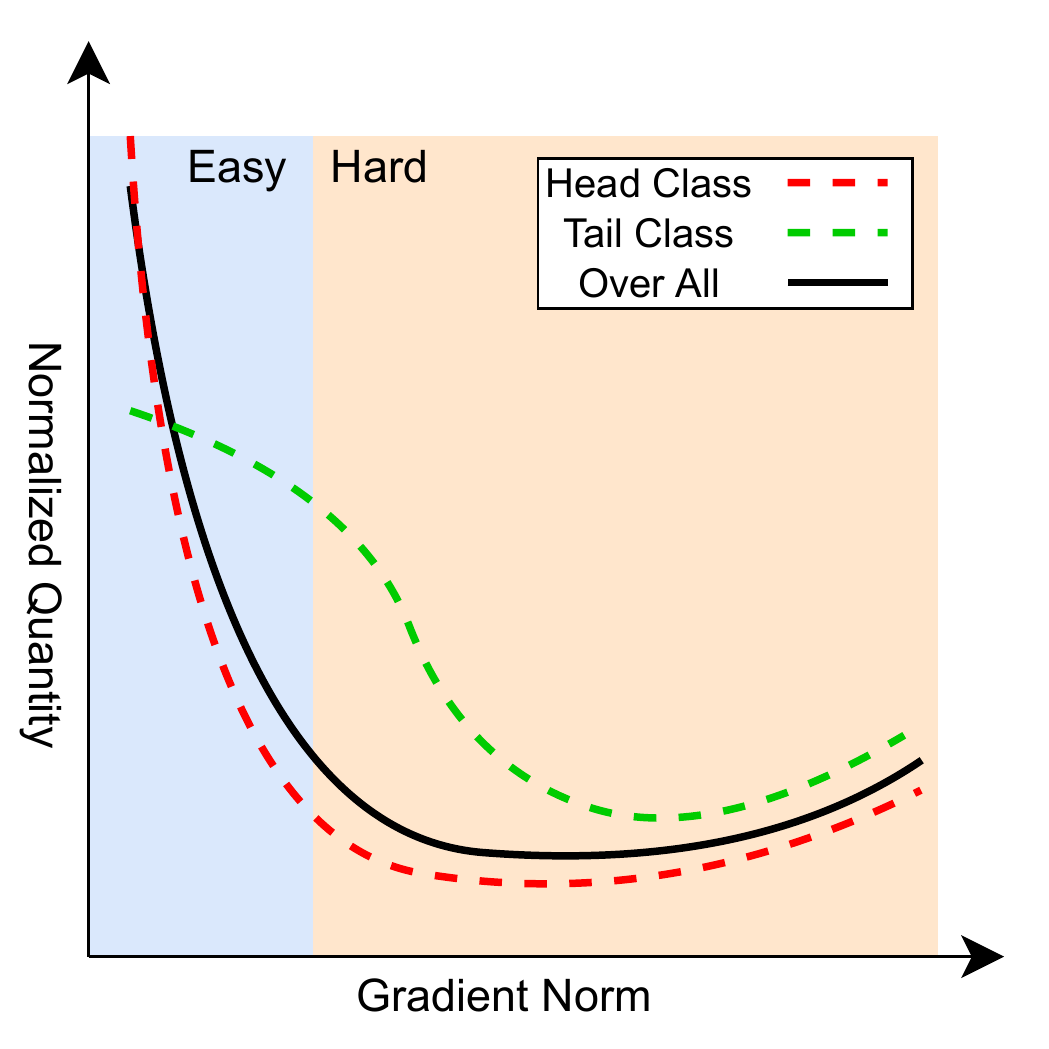}
      \label{fig:GNHs}
      % todo change the axis title 
    }
  \end{minipage}
  \begin{minipage}[b]{0.3\linewidth}
    \subfigure[URA]{
      \includegraphics[width=0.9\linewidth]{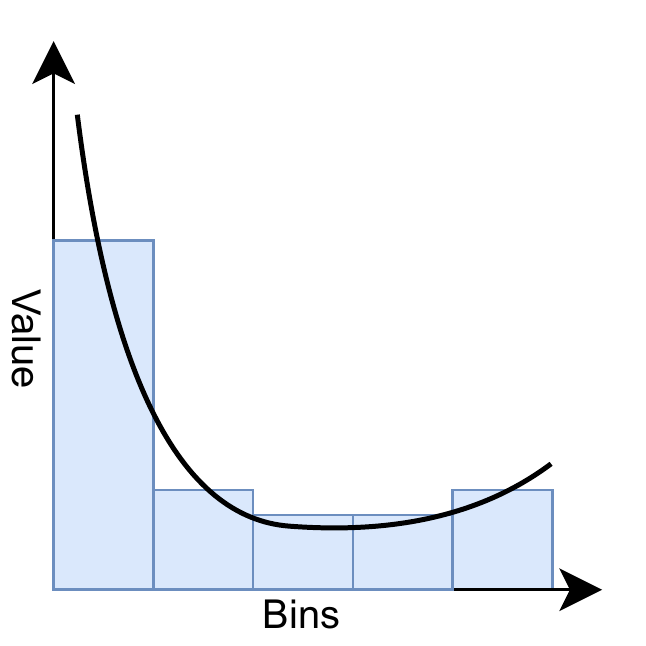}
      \label{fig:GNH}
    }
    \subfigure[AURA]{
      \includegraphics[width=0.9\linewidth]{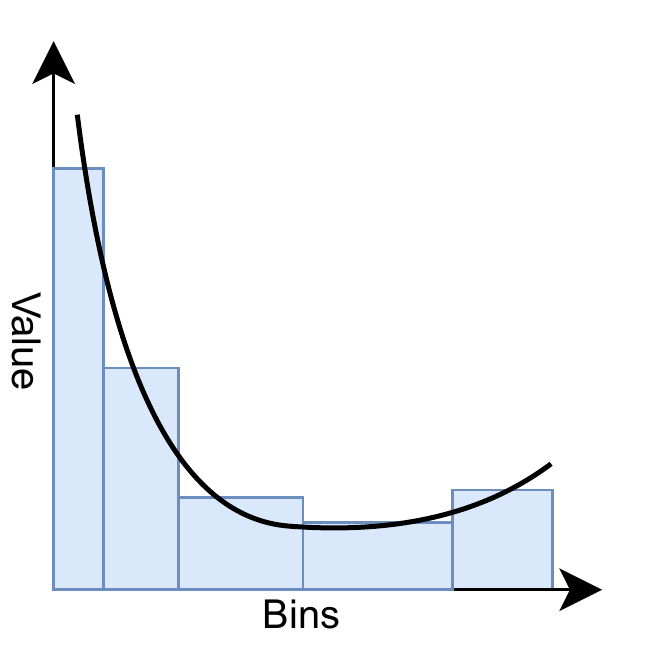}
      \label{fig:APGNH}
    }
  \end{minipage}
  \caption{ (a) Head class dominates the gradient norm distribution. 
  (b) and (c) are the approximation to the real gradient norm distribution by URA and AURA respectively. 
  The black line denotes the real distribution and the blue bins denotes the histogram.}
  \label{fig:GNHvsAPGNH}
\end{figure}
In this section, we first briefly review the Gradient harmonized mechanism (GHM) and explain its deficiency in long-tailed problems. 
Then, we introduce the proposed category-wise GHM to decouple difficulty and sample size in imbalance issues for intra- and inter-category balance strategies respectively.
For practical implementation, we further introduce the adaptive unit region approximation to attain a more precise gradient distribution for GHM without increasing the complexity, which results in performance improvement.
% }

\subsection{Preliminaries}
% \lin{
Gradient or loss are widely used as the criterion for choosing the easy and hard examples~\cite{lin2017focal, li2019gradient}. As shown in gradient norm distribution (the black line) in Fig.~\ref{fig:GNHs}, the contributions of the hard examples can be overwhelmed by the huge amount of easy examples. To solve the problem, GHM~\cite{li2019gradient} is proposed to harmonize easy and hard examples by assigning weights proportional to the reciprocal of gradient norm distribution, which is implemented by Unit Region Approximation (URA) as a trade-off between the computational complexity and performance.
However, GHM is not designed for the long-tailed problem.
%强行用会有问题，实验也证明了这一点。
As shown in Fig.~\ref{fig:GNHs},
% the gradient norm distribution of all classes (black line) would be dominated by head class (red dotted line), while the distribution of tail class (green dotted line) would be submerged
the overall gradient norm distribution (the black line) would be dominated by head class (the red dotted line), while the distribution of tail class (the green dotted line) would be submerged. 
This can be easily understood because the sample sizes of tail classes are much smaller than that of head classes.
% }
% GHM is based on gradient norm histogram (GNH), 
% which is an approximation of the real gradient norm distribution.
% In GHM, samples are divided into easy and hard samples 
% according to the gradient norm.
% As can be seen from black line in Fig.~\ref{fig:GNHs}, the contributions of the hard samples can be overwhelmed by the huge amount of easy samples.
% GHM harmonizes the contribution of easy and hard samples by assigning weight to samples according to the reciprocal of gradient norm distribution.
% Besides, the performance of GHM is related to the approximation precision of the GNH,
% and higher approximation precision would result in better performance~\cite{li2019gradient}.

\subsection{Category-Wise GHM}
Instead of calculating the overall gradient norm distribution, we propose the category-wise gradient harmonized mechanism (CWGHM) to separate the gradient norm distributions of head classes from tail classes. 
Essentially, CWGHM decouples the difficulty and sample size, which correspond to intra-category balance and inter-category balance respectively. 
The former means that CWGHM harmonizes the easy and hard of examples in each category separately, while the latter indicates that CWGHM encodes the difference in sample size among the categories, thus could be further used for inter-category balance.
% Within the histogram of each class, CWGHM harmonizes the easy and hard of examples in one category, which is the intra-category balance.
% And across the histograms of different categories,
% CWGHM reflects the difference in sample size among the categories, which could be further used for inter-category balance.

% In addition, the contribution of all examples are considered to be consistent in CWGHM.
Formally, we denote the gradient norm distribution of category $c$ as $\mathcal{H}_c$ and suppose it contains $N$ unit regions. 
In the $i$th unit region, we count the number of examples and denote it as $v_{c,i}$.
Actually, new examples are more likely to be represented by previous examples as the sample size increasing \cite{cui2019class}.
Therefore, we also introduce the effectiveness factor $\alpha$  to prevent model learning from repeated examples.
Then the effective sample size of category $c$ is defined as $S_c=\sum_{i=1}^{N}{(v_{c,i})^\alpha}$. 
To achieve the harmony of easy and hard examples via intra-category balance, the weight $W_{c,i}$ for the examples in the $i$th unit region of category $c$ is formulated as:
\begin{equation}
    W_{c,i} = \frac{1}{v^{\alpha}_{c,i}/S_c}=\frac{S_c}{v^{\alpha}_{c,i}}
    \label{eq:weight}
\end{equation}

In addition, we also deploy an asymmetric margin strategy based on the effective sample size of each category
to correct the decision boundary shift, which is the inter-category balance. 
Suppose an example from category $m$, 
$M_{m,n}$ denotes the margin added to the $n$th output logit $z_n$ of model,
which is the probability that the example belongs to category $n$.
% And $M_{m,n}$ is formulate as equation \ref{eq:margin}
\begin{equation}
  M_{m,n} = \gamma \log[ \min(1, S_n/S_m)] ,
  \label{eq:margin}
\end{equation}
where $\gamma$ is a hyperparameter to adjust the value of margin,
For the influence of head class on tail class, it will be reduced as $S_n/S_m<1$ and $M_{m,n}<0$. 
In contrast, the influence of tail classes on head classes stays the same
since $S_n/S_m>1$ and $M_{m,n}=0$.
Therefore, the head classes are protected from being influenced while the tail classes are emphasized. 

Overall, the loss function can be formulated as (\ref{eq:loss}), in which $C$ denotes all categories and $y_m$ denotes the ground truth,
\begin{equation}
\mathcal{L} = -\sum_{m\in C}{\left[ W_{m,i}\cdot y_m\log \left( \frac{e^{z_m}}{\sum_{n\in C}e^{z_n+M_{m,n}}} \right)\right]}
\label{eq:loss}
\end{equation}

\begin{table*}[t]
  \centering
  \caption{Top-1 accuracy (\%) of ResNet-50 on long-tailed ImageNet and iNaturalist2017 dataset.
  Categories are divided into four groups and
  sample size gradually decreases from ``Many'' to ``Rare''.}
  \label{tab:iamgenet}
  \resizebox{\textwidth}{!}
  {
    \begin{tabular}{l|cccc|c|ccccc}
      \toprule
      & \multicolumn{5}{c|}{Long-Tailed ImageNet(imbalance factor=100)}                                          & \multicolumn{5}{c}{iNaturalist2017(imbalance factor=435)}                                                                     \\ \cline{2-11} 
      & Many           & Medium         & Few            & Rare           & Over All       & Many           & Medium         & Few            & \multicolumn{1}{c|}{Rare}           & Over All       \\ \midrule
Softmax         & \textbf{63.91} & 48.07          & 22.15          & 12.55          & 36.67          & 84.92          & 69.88          & 50.21          & \multicolumn{1}{c|}{40.63}          & 54.64          \\
Class Balance    & 63.30          & 47.71          & 22.26          & 12.63          & 36.48          & 85.29          & 70.66          & 50.88          & \multicolumn{1}{c|}{41.31}          & 55.31          \\
Focal\cite{lin2017focal}           & 62.70          & 47.70          & 23.42          & 12.86          & 36.67          & 84.41          & 70.01          & 51.96          & \multicolumn{1}{c|}{41.13}          & 55.15          \\
EffictiveNumber\cite{cui2019class} & 63.86          & 47.91          & 23.42          & 13.70          & 37.22          & 87.77          & 72.96          & 54.45          & \multicolumn{1}{c|}{44.49}          & 58.25          \\
Seesaw\cite{wang2021seesaw}          & 63.51          & 48.69          & 24.37          & 15.52          & 38.02          & 87.48          & \textbf{73.60} & 55.64          & \multicolumn{1}{c|}{46.78}          & 59.15          \\
ours            & 63.61          & \textbf{48.73} & \textbf{25.78} & \textbf{16.95} & \textbf{38.77} & \textbf{87.84} & 73.59          & \textbf{56.75} & \multicolumn{1}{c|}{\textbf{48.02}} & \textbf{59.85} \\
\bottomrule
\end{tabular}
  }
\end{table*}

\subsection{Adaptive Unit Region Approximation}
Since the performance of GHM is dependent on the approximation precision,
the adaptive unit region approximation (AURA) is proposed to improve the approximation precision.
As shown in ﬁgure \ref{fig:GNH}, a number of unit regions (blue bins) with equal width are employed to approach the real distribution (black line) in URA. 
% The precision of GNH approximation is heavily related to the number of bins, thus reducing quantization interval would increases computational complexity. 
In fact, the real gradient norm distribution should be fitted with narrower bins at the parts of higher steepness to obtain a rather higher quantization accuracy, while wider bins are used to fit the parts of larger smoothness to reduce the computational complexity, as indicated in Fig.~\ref{fig:APGNH}.
\begin{algorithm}[t]
  \caption{Adaptive Unit Region Approximation}
  \label{alg:APGNH}
  \begin{algorithmic}
    \STATE \textbf{Initialize} $\mathcal{H}_c$ with $d_i=1/N$ (the width of unit region $i$) and $v_i=0$ (the value of unit region $i$) for $i=1,2,3,\cdots,N$;
    \FOR{each epoch}
      \FOR{each example}
        \STATE Calculate the gradient norm $g$;
        \STATE Find which unit region that $g$ lies in;
        \STATE \textbf{Update} the value of the unit region, $v_i\leftarrow v_i+1/(N*d_i)$; 
      \ENDFOR
      \STATE \textbf{Reassign} the width of each unit region, $d_i\leftarrow 1/\log(v_i)$;
      \STATE Normalize the width of all unit region to 1;
      \STATE Reset $v_{c,i}$ to 0;
    \ENDFOR
  \end{algorithmic}
\end{algorithm}
% For each category, a corresponding adaptive precision GNH with $N$ bins, $\mathcal{H}_c$, is applied to count the gradient norm distribution of category $c$.
Given gradient norm distribution $\mathcal{H}_c$, we denote the width and value of the $i$th unit region in $\mathcal{H}_c$ as $d_{c,i}$ and $v_{c,i}$ respectively.  
Then $\mathcal{H}_c$ is updated following Algorithm \ref{alg:APGNH}.
All unit regions are initialized with the same width $d_{c,i}=1/N$ and $v_{c,i}=0$.
Then $\mathcal{H}_c$ will count the gradient norm of every example during previous epoch.
After each epoch, the width of each unit region is reassigned according to the gradient norm distribution counted in the last epoch.

With AURA, we can approximate the gradient norm distribution with a higher precision, which result in the performance improvement of the proposed method.

\section{Experiments}
\subsection{Experimental Settings}
% \subsubsection{Datasets}
Our experiments are performed on CIFAR \cite{krizhevsky2009learning, torralba200880}, Tiny-ImageNet$\footnote{The dataset can be downloaded from https://www.kaggle.com/c/thu-deep-learning/data}$, ImageNet ILSVRC 2012 \cite{deng2009imagenet, russakovsky2015imagenet} and iNaturalist dataset. 
Following \cite{cui2019class}, long-tailed versions of these datasets 
with varied imbalance factors, 
which means the ratio of the sample size of head classes to tail classes, 
are constructed by randomly removing examples.
Besides, iNaturalist dataset is a long-tailed large-scale real-world dataset, whose imbalance factor is 435.
With the above datasets, extensive studies are conducted to demonstrate the effectiveness of the proposed method. 
For our method, the best hyper-parameter is chosen via cross-validation, $\alpha=0.9$. 
Besides, $N=30$ and $\gamma=0.8$, which are following \cite{li2019gradient} and \cite{wang2021seesaw}.
For other methods, we use the hyper-parameters provided by the authors. 

% \subsubsection{Implementation Details}
% Experiments on TinyImageNet-LT and CIFAR-LT are implemented with ResNet-32 \cite{he2016deep}. 
% Besides, ResNet-50 \cite{he2016deep} is implemented on ImageNet-LT, whose imbalance factor is 100, and iNaturalist2017 dataset
% The model is trained with batch size of 128 for 200 epochs. 
% The learning rate is 0.1 and it will decrease by 0.01 after 160 and 180 epochs, respectively.
% , and trained with batch size of 1024 for 100 epochs. The learning rate is 0.1 and it will decrease by 0.1 every 30 epochs.

\begin{table}
  \caption{Top-1 accuracy (\%) of ResNet-50 on long-tailed ImageNet. 
   ``R" represents intra-category balance, ``A$^{URA}$" and ``A$^{AURA}$" represent inter-category balance implemented by URA and AURA respectively. }
  \label{tab:ablation1}
  \resizebox{\linewidth}{!}
  {
    \begin{tabular}{l|cccc|c}
      \toprule 
                      & \multicolumn{5}{c}{ImageNet-LT(imbalance factor=100)}          \\ \cline{2-6} 
                      & Many & Medium & Few & Rare  & Over All \\
                      \midrule 
    Softmax         & 63.91    & 48.07  & 22.15      & 12.55 & 36.67    \\
    R           & 63.51    & 48.69  & 24.37      & 15.52 & 38.02    \\
    A$^{URA}$        & 63.93    & 48.32  & 22.60       & 13.44 & 37.07    \\
    R+A$^{URA}$  & 63.55    & 48.67  & 25.16      & 16.30  & 38.42    \\
    R+A$^{AURA}$ & 63.61    & 48.73  & 25.78      & 16.95 & \textbf{38.77}   \\
    \bottomrule
    \end{tabular}
  }
  \end{table}

\subsection{Experimental Results}
Extensive studies on long-tailed CIFAR-10/100 and TinyImageNet datasets with varied imbalance factors are conducted with ResNet-32.
Table \ref{tab:big_table} shows the performance of ResNet-32 using various methods in terms of classification accuracy.
We present results of using softmax cross-entropy loss, class balance loss, which assigns weights to each classes according to the inverse of sample size, focal loss \cite{lin2017focal}, GHMc loss \cite{li2019gradient}, effective number loss \cite{cui2019class}, Equalization loss \cite{tan2020equalization}, Seesaw loss \cite{wang2021seesaw}, and our proposed method.

From the results in Table \ref{tab:big_table}, we have the following observations:
(1) Our method achieves the best performance on datasets with large imbalance factors.
(2) For balanced datasets, whose imbalance factor is 1, our method has almost the same performance as softmax cross-entropy loss.
(3) The original form of GHM is not suitable for the multi-class classification task.

% In order to validate that our method on tail classes, 
% the accuracy of long-tailed CIFAR-10 dataset is shown in Fig.~\ref{fig:compare}.
% The classes are divided into four groups according to the sample size, which gradually decreases from ``Many" to ``Rare", and corrsponds to the classes from head classes to tail ones.
% It can be seen that our method has a significant improvement on tail classes,
% and it keeps the comparable accuracy with other methods on head classes. 
% This verifies the effectiveness of our method on long-tailed datasets.

We also present results on 
ImageNet-LT and iNaturalist2017 dataset 
to demonstrate the performance on large-scale real-world datasets.
Table \ref{tab:iamgenet} summarizes the top-1 accuracy 
on ImageNet-LT and iNaturalist2017 datasets. 
Our method achieves the best performance on large-scale real-world datasets, and the gain mainly comes from tail classes while keeping the performance on the head class without sacrificing head classes. 
% \lin{emphasizing improving the tail classes significantly  but not saying the improvement comes from tail class.}
Still, our method obtains a more significant gain on iNaturalist2017 than ImageNet-LT, even with a larger imbalance factor. 
% Abundant Common Frequent Occasional Rare

\subsection{Ablation Studies}

% the effect of inter- and intra-category balance 

Table \ref{tab:ablation1} shows the ablation study of 
each module of the proposed method on Long-Tailed ImageNet.
It can be seen that each module can improve the performance 
separately.
Intra-category balance enhances the classification performance 
on both head classes and tail classes.
As a complement, inter-category balance imposes a significant improvement 
on the tail classes. 
Furthermore, the proposed AURA achieves superior performance to URA 
as AURA provides a better approximation to the real distribution of gradient norm distribution.
Based on all these modules, our method achieves the overall best performance.

\section{Conclusion}
In this paper, we propose a novel
gradient harmonized mechanism with category-wise adaptive precision to decouple the difficulty and sample size in long-tailed problems, corresponding to the intra- and inter-category balance strategies.
Intra-category balance emphasizes hard examples in each category to optimize the decision boundary, while inter-category balance aims to correct the shift of decision boundary by regarding all examples of one category as a unit.
Besides, adaptive unit region approximation is proposed to improve the performance further.
% In this work, we have presented a approach to tackle the long-tailed problem in real-word dataset.
% The key idea is emphasizing the contribution of hard samples to obtain a optimal decision space.
% Therefore, we solve the long-tailed problem by combining intra-category balance and inter-category balance. 
% % Following this idea, we further propose a class-balanced loss to re-weight loss inversely with the effective number of samples per class. 
% Extensive experiments have been conducted to understand and analyze the proposed method.
% % The proposed method has been verified by experiments on both CIFAR and large-scale datasets ImageNet.
It demonstrates that the proposed method achieves the comprehensive balance for the long-tailed problem and outperforms other approaches on all datasets.
% \cleardoublepage
% References should be produced using the bibtex program from suitable
% BiBTeX files (here: strings, refs, manuals). The IEEEbib.bst bibliography
% style file from IEEE produces unsorted bibliography list.
% -------------------------------------------------------------------------
\bibliographystyle{IEEEbib}
\bibliography{refs}

\end{document}